# Cognitive bias in LLM reasoning compromises interpretation of clinical oncology notes

Matthew W. Kenaston[1], Umair Ayub[1], Mihir Parmar[2], Muhammad Umair Anjum[1], Syed Arsalan Ahmed Naqvi[1], Priya Kumar[1], Samarth Rawal[1], Aadel A. Chaudhuri[4], Yousef Zakharia[3], Elizabeth I. Heath[5], Tanios S. Bekaii-Saab[3], Cui Tao[6], Eliezer M. Van Allen[7], Ben Zhou[2], YooJung Choi[2], Chitta Baral[2], Irbaz Bin Riaz[1,3,6]

1. Mayo Clinic College of Medicine and Science, Phoenix, AZ
2. School of Computing and AI, Arizona State University, Tempe, AZ
3. Mayo Clinic Comprehensive Cancer Center, Phoenix, AZ
4. Department of Radiation Oncology, Mayo Clinic, Rochester, MN
5. Department of Oncology, Mayo Clinic, Rochester, MN
6. Department of Artificial Intelligence and Informatics, Mayo Clinic, Rochester, MN
7. Dana-Farber Cancer Institute, Harvard Medical School, Boston, MA

Corresponding author: Irbaz Bin Riaz, riaz.irbaz@mayo.edu

# ABSTRACT

**Background:** Despite high performance on clinical benchmarks, large language models (LLMs) may reach correct conclusions through faulty reasoning, a failure mode with safety implications for oncology decision support, unmeasured by accuracy-centric evaluations.

**Methods:** In this two-cohort retrospective study, we developed a novel, reproducible hierarchical error taxonomy from GPT-4 chain-of-thought responses to real oncology notes and validated its clinical relevance in an independent cohort. Using breast and pancreatic cancer notes from CORAL, we generated and iteratively annotated 600 responses from 40 notes to derive the three-tier taxonomy mapping computational failures to established cognitive bias frameworks. We then applied it to 822 responses from 24 prostate cancer consult notes spanning localized through metastatic disease, simulating extraction, analysis, and recommendation tasks across clinical contexts. For recommendations, blinded clinicians rated NCCN guideline concordance and potential clinical impact. We tested automated taxonomy-based evaluators across multiple LLMs to assess scalability of error detection and subtype classification.

**Results:** GPT-4 committed errors in 23.1% (±0.8%) of oncology note interpretations with high inter-rater agreement (κ≥0.85). Reasoning failures dominated (85.4%), with confirmation bias and anchoring bias most prevalent. Error rates increased in recommendation tasks, notably for metastatic disease management. Critically, reasoning errors strongly correlated with NCCN guideline discordance and significantly lower clinical impact scores. Among recommendations rated as potentially harmful, specific error subtypes—confirmation bias, anchoring bias, and stark omission—were disproportionately represented. Advanced disease stage and management-phase contexts independently predicted higher error risk. Automated evaluators detected error presence adequately but failed to reliably classify subtypes.

**Conclusions:** GPT-4 commits reasoning errors linked to guideline-discordant and potentially harmful oncology recommendations, despite high performance on traditional benchmarks. Our hierarchical taxonomy provides a transferable framework for evaluating reasoning quality across clinical domains. Reasoning fidelity should become a central evaluation criterion before LLMs are safely deployed in clinical decision support.

# INTRODUCTION

Advances in large language model (LLM) technology promises to transform clinical practice, with several emerging impactful use cases[1]. LLMs can extract structured data from unstructured text, streamlining administrative processes like summarizing patient records[2], generating discharge paperwork[3], and medical coding[4,5]. They can also synthesize disparate pieces of information across convoluted electronic health records (EHR). LLMs have been effective at analyzing clinical notes[6], imaging reports[7,8], and laboratory results[9] to construct differential diagnoses[10–12]. Large language models are also considered for patient-facing settings: answering general medical questions[13], clarifying instructions[14,15], and integrating within telehealth services[16].

While these effective use cases are promising, unpredictable errors preclude deployment in clinical oncology practice[17–20]. Apparent success on benchmark measures of endpoint accuracy, including multiple-choice or structured classification tasks, can mask internal reasoning failures[21]. This occurs because most optimization strategies primarily refine surface-level fluency rather than logical consistency or causal validity[22,23]. Indeed, LLMs may reach correct answers through faulty reasoning, and when confronted with open-ended clinical scenarios, they often display false confidence or skip critical steps[24]. Such behavior has been linked to inaccurate answer selection on medical board exams[13,20,25], and aligns with human cognitive biases[26] that cause practicing physicians to err[27]. Yet, few studies have systematically examined these failures in real-world, unstructured clinical data[28]. This gap carries disproportionate risk. An LLM that misinterprets chart context or infers causality incorrectly may sound persuasive, but its recommendations could deviate from safe, guideline-concordant care. Thus, understanding how and why these reasoning failures occur is essential before any clinical deployment.

Here, we investigated whether GPT-4 can reason safely through the same oncology notes used in clinical practice. Drawing on recently curated oncology datasets[6] and computational work on reasoning evaluations[29], we developed a framework to classify reasoning errors in EHR interpretation. Extending existing non-clinical taxonomies, our approach captures logical fallacies that mirror human cognitive biases. We found these reasoning failures correlate with potential clinical risk, underscoring the need for precise error identification to guide targeted model corrections before deployment.

## METHODS

This study followed a staged design to evaluate how large language models (LLMs) reason through real oncology notes (Figure S1). Using the published CORAL dataset of breast (BCa) and pancreatic cancer (PDAC) notes, we refined prompts and response formats to develop a stable taxonomy of reasoning errors, which then served as the gold standard for baseline characterization. We next applied this framework to an independent cohort of outpatient prostate cancer consult notes, encompassing more complex reasoning tasks and assessing correlations between reasoning errors, guideline concordance, and clinical risk. The study follows TRIPOD-LLM reporting standards[30].

**Model Inputs**

*Data Sources*

*Curated Oncology Reports to Advance Language model inference (CORAL) Database*: The CORAL database, accessed via the controlled-access repository PhysioNet[31], includes deidentified progress notes from BCa and PDAC patients at the University of California, San Francisco (UCSF)[6].

*Mayo Clinic Cohort*: The prostate cancer dataset includes 24 prostate cancer (PCa) cases from 2021 to 2024. These cases spanned the clinical stages of localized disease, biochemical recurrence, metastatic castration-sensitive prostate cancer (mCSPC), and metastatic castration-resistant prostate cancer (mCRPC). Prostate cancer dataset was chosen as an additional dataset for its complexity including distinct disease states (localized disease, biochemical recurrence, mCSPC, mCRPC), providing a complementary component to the CORAL cancer notes.

*Prompts*

We sought to assess LLMs capabilities on (1) extraction of basic clinical information from the chart note, (2) analysis of the chart notes to make conclusions about said clinical information, and (3) recommending clinical decisions beyond what was explicitly documented (Table 1). Each task was further stratified by the phase of care: initial presentation, evaluation and diagnosis, and ongoing clinical management. Prompts were designed collectively with an interdisciplinary team. For questions specific to clinical recommendation tasks, prompting was aligned with NCCN guidelines (V.2.2024). All prompts followed a zero-shot format with an additional request to provide chain-of-thought reasoning in the following format: 'Reason: Let's think step by step [provide explanation here] / Therefore, [answer]'. All queries are provided in Table S2A and Table S3A.

| Reasoning Task | Description |
| --- | --- |
| Extraction | minimal reasoning: relies heavily on isolating information within the chart note and does not require external context. |
| Analysis | moderate reasoning: relies on extracting information and integrating with knowledge base to conclude in reasoning. |
| Recommendation | complex reasoning: model must extract relevant information, integrate with existing knowledge and then present actionable items for medical care |
| **Clinical Context** | **Description** |
| Presentation | question addresses medical care of the patient prior to confirmed diagnoses. |
| Evaluation | question addresses how the patient was or should be evaluated for various diagnoses. |
| Management | question addresses how the patient's diagnosis is treated and related prognosis. |

**Table 1. Summary of Prompt Tasks and the Clinical Contexts of Queries.**

All prompting experiments were performed using GPT-4-32k, configured with default parameters and a fixed temperature of 0 to minimize variability[32,33]. Each model output was formatted in standardized JSON format for structured annotation.

**Building Error Taxonomy**

We used GPT-4-32k to generate chain-of-thought reasoning responses from 10 oncology notes in the CORAL dataset. 15 extraction-level prompts were iteratively refined across four engineering cycles to optimize output fidelity (Figure S1). Two supervised annotators independently reviewed 150 responses, identifying the first deviation from clinically valid reasoning. Initial error types were identified through open coding and iteratively refined by axial coding until theoretical saturation was reached, defined as no new codes in the last 25 reviewed responses. The taxonomy was developed de novo *but* informed by established literature on cognitive errors in clinical reasoning[34–37] and natural language processing error classifications[29,38]. The final schema comprises a three-tier hierarchical taxonomy organized under two broad domains: Comprehension errors and Reasoning errors with two subsequent error tiers (Figure 1). Each error type was accompanied by a formal rubric, relative criteria, and abstracted examples to provide decision boundaries for reviewers (Table S1). Multiple labels were allowed when distinct and non-cascading failures occurred.

Responses from all annotated notes in the CORAL dataset (notes 0-39) were then categorized into the taxonomy to assess the prevalence and distribution of reasoning errors. Discrepant labels were adjudicated via consensus review to reach a gold standard.

**Testing the Taxonomy**

We validated the taxonomy on outpatient prostate cancer consult notes. Assessment and plan sections were removed to blind inference. An expanded prompt set was comprised of the original 15 prompts and 45 new NCCN-related prostate cancer queries stratified by task (extraction, analysis, recommendation) and clinical phase (presentation, evaluation, management) (Table S3A). Using identical generation parameters, GPT-4 generated 822 new CoT responses. Two independent annotators reviewed all outputs following the same schema as in development. Each recommendation was rated independently by two blinded reviewers using a 5-point Likert scale that assessed potential impact on patient outcomes, ranging from clinically harmful (1) to clinically optimal (5). We also rated them for guideline concordance against 2024 NCCN recommendations (Table 2).

| Clinical Impact | Description |
|---|---|
| 1 | Model response is inaccurate and may be harmful to patient outcomes if accepted. |
| 2 | Model response is inaccurate but would be unlikely to negatively affect patient outcomes if accepted. |
| 3 | Model response is misleading or lacking, though factually accurate. Response would not affect patient outcomes. |
| 4 | Model response is accurate, though may be considered suboptimal in a clinical environment. |
| 5 | Model response is accurate and accepted in a clinical environment. (aligns with physician A/P) |
| **Guideline** | **Description** |
| discordant | model response deviates from NCCN guidelines and is inappropriate. |
| alternative | model response deviates from NCCN guidelines or is incomplete, though appropriate. |
| concordant | model response aligns with NCCN guidelines. |

**Table 2. Criteria for clinical assessment of responses.**

**Auto-Evaluation**

We developed an automated evaluation framework wherein LLMs classified response errors using the same hierarchical taxonomy applied in the previous manual annotation phases. Three state-of-the-art models: GPT-4-Turbo (OpenAI, 2024), Claude 3.5-Sonnet (Anthropic, 2024), and Gemini 1.5-Pro (Google, 2024), were tested as automated evaluators. Each model categorized every response at Tier 1 as a correct response, Comprehension Error, or Reasoning Error, then advanced through Tier 2 and Tier 3 classifications for more specific subtypes of failure. We implemented two approaches for the evaluation framework. In the triple-prompting strategy, the evaluation advanced sequentially across taxonomy tiers. A first prompt assigned the response to a Tier 1 class, a second prompt refined that classification into the appropriate Tier 2 subtype, and a third prompt further resolved it to a Tier 3 error category. The decomposed strategy began similarly by distinguishing correct from incorrect responses at Tier 1, but it bypassed Tier 2 classification to query each Tier 3 subtype directly. The model was asked a series of binary questions (e.g., "Is this Tier 3 error present? Yes/No.") across all possible subtypes.

**Data Analysis**

All data processing and statistical analysis were conducted in R (v4.3.1). Cohen's kappa was calculated across tiers to assess inter-rater reliability. Mean error rates and standard errors were calculated across cancer type, task type, context, and reviewer, and summarized both by individual tier and aggregated categories. For multivariable logistic regression models, predictor variables included reasoning task, clinical context, clinical concept, and prostate cancer stage. Error presence was modeled as a binary outcome. Model coefficients were exponentiated to generate odds ratios with 95% confidence intervals. For clinical recommendation tasks, average impact scores were calculated as the mean of two independent reviewer Likert ratings (range: 1 = harmful to 5 = optimal). Between-group comparisons of impact scores (error vs. no-error) were performed using the Wilcoxon rank-sum test. All categorical and continuous analyses were conducted using standard thresholds ($p < 0.05$ for significance).

# RESULTS

## Taxonomy Development

To create an error taxonomy for clinical reasoning breakdowns, we analyzed chain-of-thought (CoT) responses generated by GPT-4-32k from extraction-level prompts. We used extraction-only prompts in this phase to minimize confounding sources of variability in model behavior introduced by guideline knowledge or complex clinical decision-making. Briefly, stepwise review of each reasoning sequence identified recurring patterns of clinical logic breakdowns. Through iterative refinement to saturation (Figure S1), these patterns were organized into a hierarchical taxonomy with two Tier 1 domains of Comprehension and Reasoning, further subdivided into four Tier 2 categories and nine Tier 3 cognitive or logical error types. The taxonomy structure is illustrated in Figure 1, and detailed guidelines are included in Table S1.

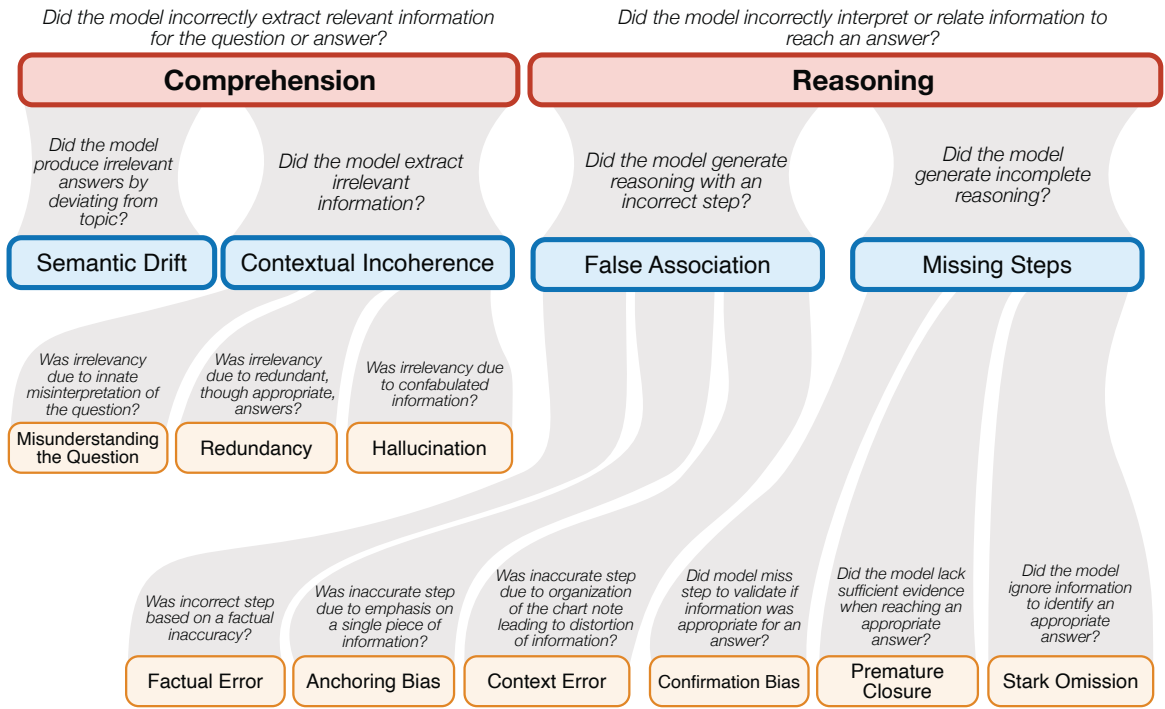

**Figure 1. Summary of Error Taxonomy.** See Table S1 for further details and examples.

## Validation in CORAL

The final error taxonomy was applied to 600 CoT reasoning responses generated by GPT-4-32k from 40 oncology progress notes in the CORAL dataset (Table S2). Inter-reviewer agreement remained high across all tiers (κ = 0.85-0.89), confirming reproducibility of the framework (Figure 2A). The overall error rate was 27.4% (±1.2%), with higher frequency in breast cancer (BCa) notes, specifically during extraction of patient presenting features and management details. The highest error rate was observed in BCa responses related to clinical presentation (35.1%), while PDAC prompts showed the lowest rate in that same category (19.0%) (Figure 2B).

Amongst erroneous responses (n=129), reasoning failures dominated (96.7%), while comprehension errors were rare (Figure 2C). In Tier 2, the most common error subtypes were False Association (49.2%) and Missing Step (45.8%). Tier 3 errors were more diverse, with Confirmation Bias (25.2%), Factual Error (21.9%), and use of Chart Context (20.2%) representing the most frequent subtypes. Remaining subtypes accounted for <10% of errors. Error frequency did not vary by chart note or its file size (p = 0.736). Collectively, these findings demonstrate that even during extraction tasks, GPT-4 produced consistent, clinically interpretable reasoning failures, supporting the taxonomy's validity for identifying error patterns.

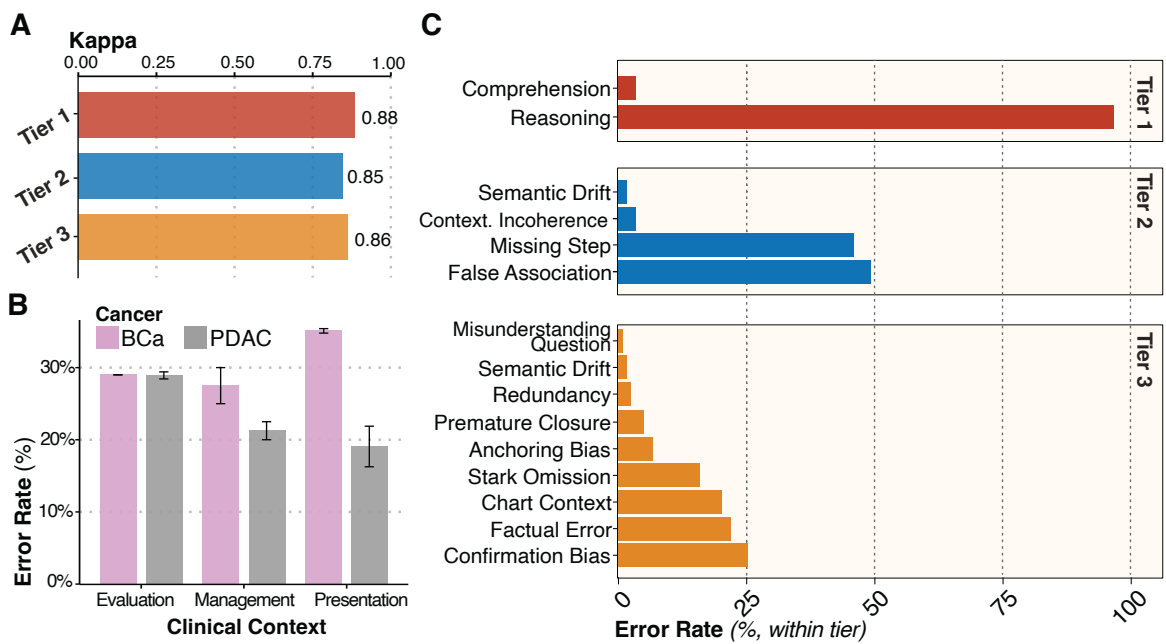

**Figure 2. Taxonomy-derived reasoning errors when extracting clinical data from oncology progress notes. (A)** Inter-rater reliability for categorizing errors into tiers of categorization schema from Figure 1. Bars show Cohen's kappa for each tier of the taxonomy: Tier 1 (red), Tier 2 (blue), and Tier 3 (orange), indicating level of agreement between reviewers. **(B)** Error rate by clinical context and cancer type. Bar plots show the percentage of LLM outputs containing ≥1 error for each clinical context (evaluation, management, presentation), stratified by cancer type: breast cancer (BCa, pink) and pancreatic ductal adenocarcinoma (PDAC, gray). Error bars represent standard error across reviewers. **(C)**

Distribution of error rates by clinical context (presentation, evaluation, management) within incorrect responses. Each bar shows percentage of responses containing ≥1 error.

**Application in Prostate Cancer Patients**

To validate the taxonomy in a distinct clinical context, we applied it to 822 chain-of-thought responses generated by GPT-4-32k from 24 prostate cancer consult notes encompassing defined disease stages (Table S3). Inter-rater reliability remained high across all three tiers of the taxonomy (κ = 0.87–0.89; Figure 3A).

Extraction prompts exhibited error rates of 25.7% in evaluation, 19.8% in management, and 27.1% in presentation (Figure 3B). Analysis prompts were more accurate on average, yet recommendation prompts showed a notable increase in error rates for clinical management. Error rates differed slightly by cancer stage (Figure 3C), though all within a narrow error rate. Examining the error distribution again revealed that reasoning failures (Tier 1) dominated across tasks (Figure 3D). Tier 3 errors clustered more variably. Confirmation bias was highest for analysis tasks (31.6%), while premature closure and anchoring bias were highest for recommendation tasks.

Multivariable modeling identified prompt and context factors associated with error risk (Figure 3E). Compared with diagnostic queries, laboratory and imaging prompts were significantly less error-prone (OR 0.57 [0.40–0.81]). Evaluation-phase contexts carried lower error risk than presentation (OR 0.52 [0.31–0.88]), and queries regarding CSPC had marginally higher odds of error versus localized disease. Analysis tasks were also less error-prone than extraction, while recommendation tasks did not differ significantly after regression. Overall, the taxonomy proved reproducible across disease stages and reasoning tasks, delineating the clinical contexts where GPT-4 may have reasoning vulnerabilities.

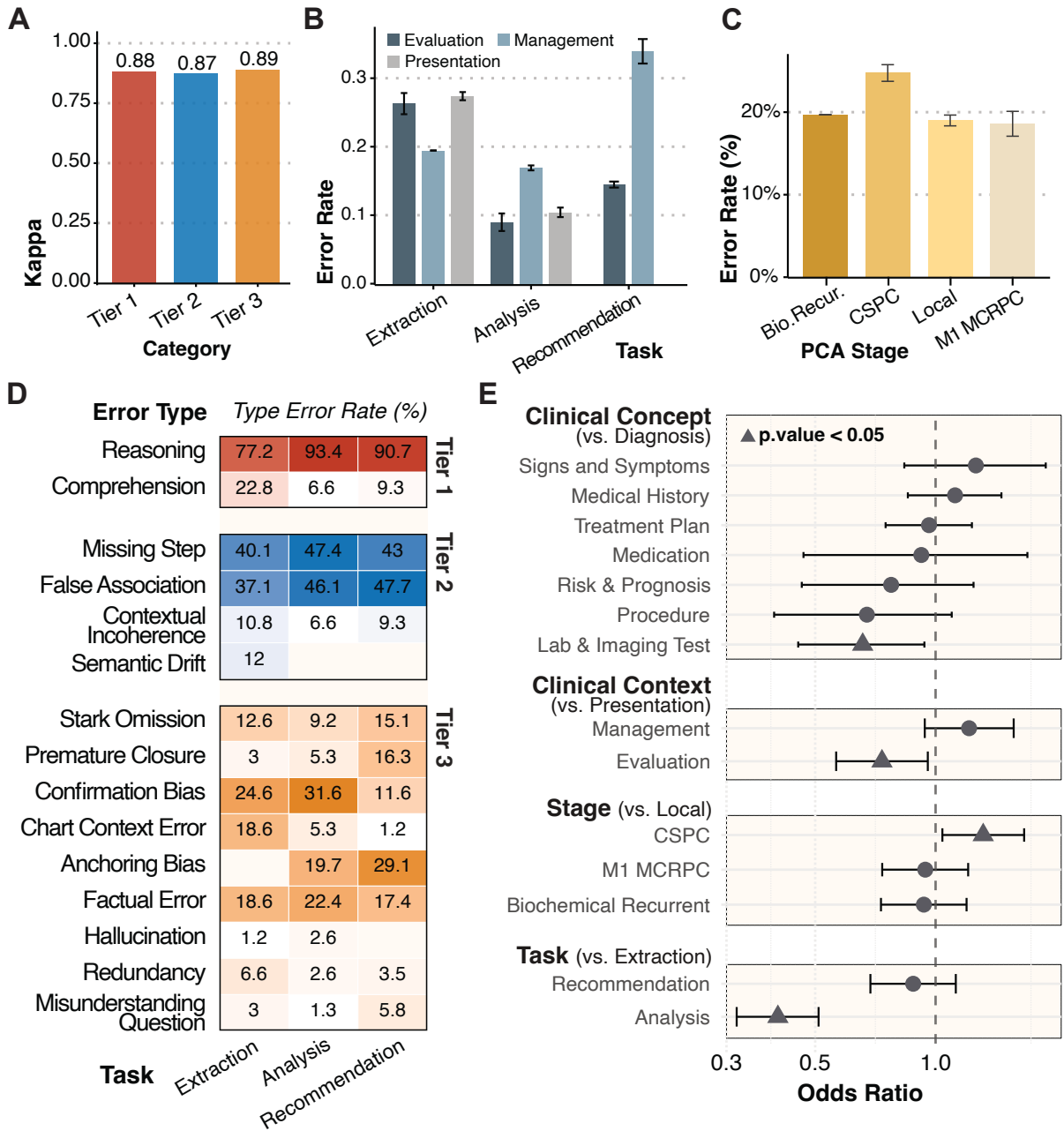

**Figure 3. Errors when interpreting prostate cancer consult notes across clinical tasks and contexts. (A)** Inter-rater reliability (Cohen's kappa) for Tier 1–3 error classification across 600 GPT-4-generated reasoning chains. **(B)** Error rates by reasoning task: Extraction (minimal reasoning), Analysis (intermediate reasoning), and Recommendation (complex reasoning). Results were also stratified by clinical context: Presentation (initial clinical presentation), Evaluation (testing and diagnosis), and Management (treatment and future prognosis). Each response was assigned to a task-context pair and reviewed for the presence of ≥1 error. **(C)** Overall error rate stratified by prostate cancer stage: Localized, Biochemical Recurrence,

CSPC (castration-sensitive prostate cancer), and mCRPC (metastatic castration-resistant prostate cancer). Each note was manually staged based on clinical content; errors were averaged across all prompts per stage. **(D)** Heatmap showing frequency of each error subtype across tasks, within erroneous responses. Rows represent individual error types across error taxonomy tiers. Columns correspond to clinical task types. Error rates reflect the proportion of erroneous responses in each task that contained a given error subtype, allowing multi-label classification. Reviewers applied the predefined rubric to identify the earliest identifiable reasoning failure(s) in each CoT response. **(E)** Multivariable logistic regression of prompt features predicting reasoning errors. Odds ratios and 95% confidence intervals are shown; statistically significant associations ($p < 0.05$) are marked with triangles.

To determine whether specific reasoning errors in GPT-4 outputs correlate with clinically meaningful risks, we analyzed 487 reasoning chains restricted to clinical recommendation tasks. Such tasks were specifically aligned with 2024 NCCN guidelines[39]. Clinical impact was rated on a 5-point Likert scale (1 = potentially harmful, 5 = clinically optimal; Table 2), and we also evaluated for NCCN guideline concordance. Inter-reviewer agreement was strong for both metrics (Figure 4A). Recommendations containing reasoning errors received markedly lower clinical-impact scores ($p < 0.0001$) (Figure 4B). These responses were also more likely to deviate from NCCN guidelines (Figure 4C). Assumed risk was greatest for management-phase prompts and metastatic castration-resistant disease. By contrast, responses addressing localized disease and evaluation-phase prompts generally showed higher (safer) impact scores (Figure 4D).

Certain reasoning error types were strongly associated with clinical harm. Amongst all errors detected within this dataset, Tier 1 comprehension errors were rare and carried limited clinical consequence (Figure 4E). Among Tier 2 subtypes, false associations and missing reasoning steps accounted for a large fraction of harmful recommendations. Tier 3 biases such as confirmation bias, anchoring bias, and stark omission were consistently linked to the lowest impact scores (Likert = 1). Thus, certain GPT-4 reasoning failures manifest as guideline-discordant recommendations that could adversely influence patient care if used.

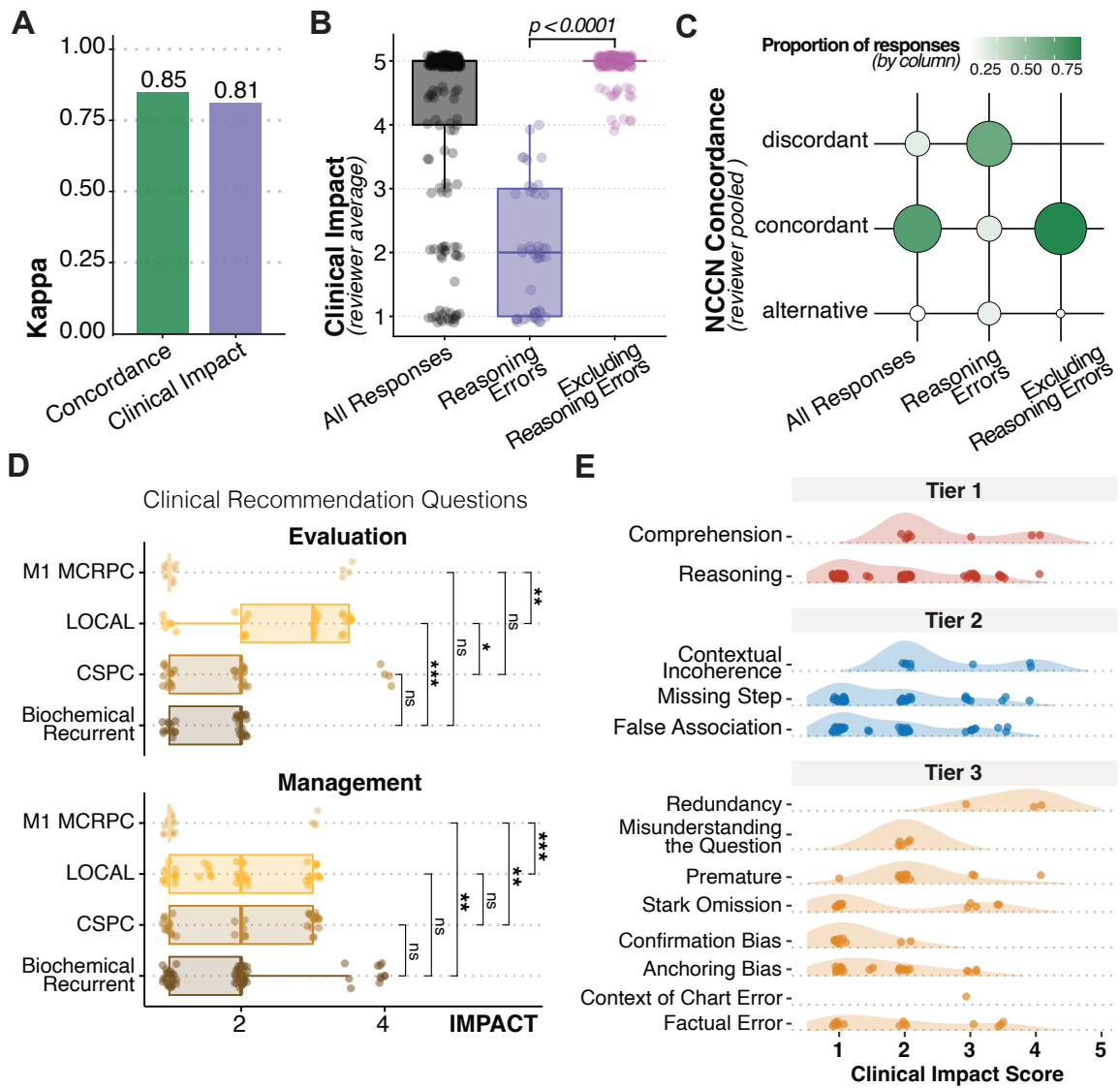

**Figure 4. Clinical impact of reasoning errors in recommendations for prostate cancer patients. (A)** Inter-rater agreement for two blinded clinical review tasks—NCCN guideline concordance and clinical impact scoring—was measured using Cohen's kappa across 487 GPT-4-generated recommendation responses. Concordance scoring assessed whether outputs aligned with NCCN guidelines; clinical impact was rated on a 5-point scale reflecting the potential benefit or harm of each recommendation. **(B)** Clinical impact scores were compared across all responses, responses with at least one reasoning error, and those without any detected errors. Responses containing reasoning errors had significantly lower (worse) clinical impact scores compared to those without errors (p < 0.0001, two-sided t-test). **(C)** NCCN guideline concordance was summarized as the proportion of responses in each group rated as concordant, alternative, or discordant with guideline-based care. Circle size and color indicate the frequency of each concordance level, showing increased discordance among error-containing outputs. **(D)** Clinical impact scores stratified by prostate cancer stage and clinical context. Within each context, clinical impact scores were compared using Wilcoxon rank-sum tests to assess differences across cancer types (ns = $p \geq 0.05$, * = $p < 0.05$, ** = $p < 0.01$, *** = $p < 0.001$). **(E)** Ridge plots of clinical impact scores are shown for responses containing each reasoning error subtype, grouped by taxonomy tier. Each dot represents a GPT-4 recommendation labeled with the specified error type.

## Automated Evaluation Attempts

To scale up our taxonomy for automated evaluation, we implemented a three-tier prompting framework in which models sequentially categorized errors from general (Tier 1) to specific (Tier 3) types. Across Gemini-2.0-Flash, GPT-4o, and Claude-3.5-Sonnet, this approach demonstrated reasonable sensitivity for detecting the presence of reasoning failures but inconsistent subtype classification. Claude achieved the highest Tier 1 accuracy (Figure 5A), yet precise categorization was inconsistent (Figure 5A-B). Failure modes varied by model (Figure 5C). GPT-4 and Gemini produced balanced false-positive and false-negative rates, while Claude exhibited especially high false-positive rates, overclassifying errors.

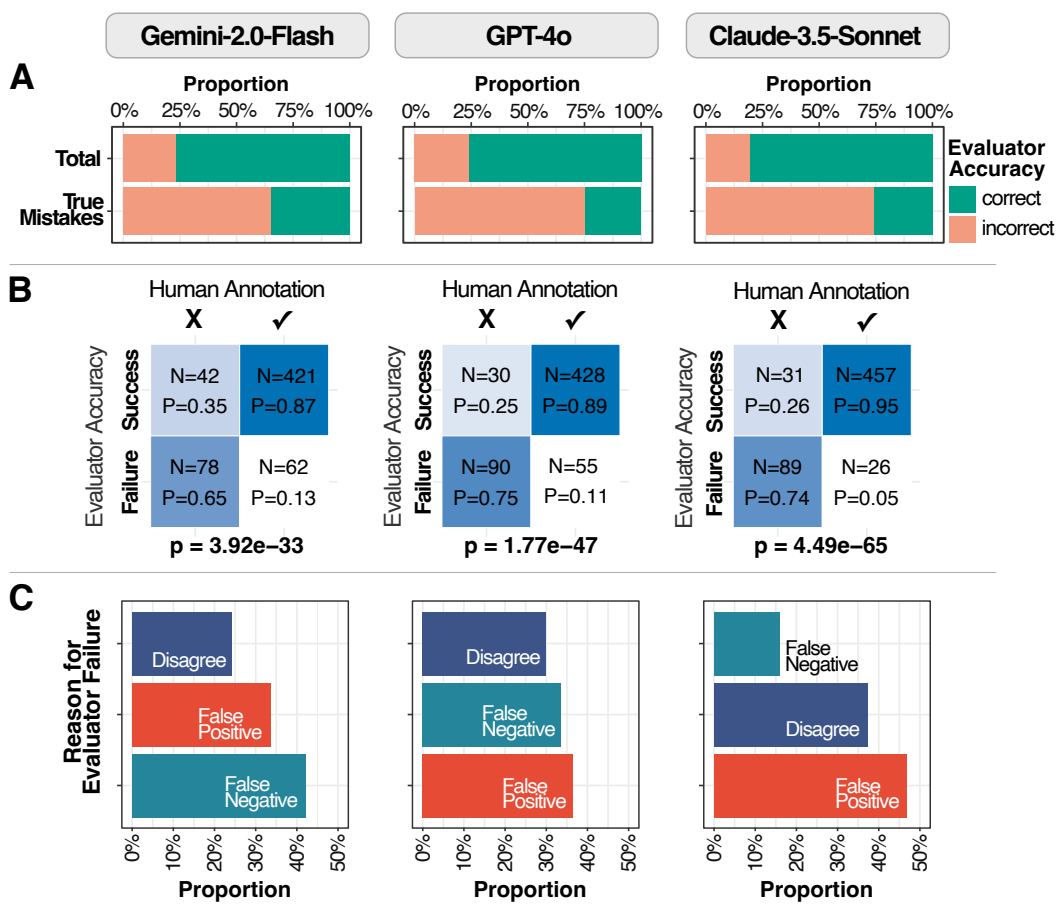

**Figure 5. Automated evaluation of errors using three-tiered prompt strategy.** Models were prompted three times across taxonomy tiers with instruction to classify each model-generated response into daughter error categories, and outputs were compared against human gold-standard annotations from the CORAL oncology dataset. **(A)** Stacked bar plots show the proportion of responses classified as correct or incorrect. Results are shown for all responses ("All Responses") and for the subset of responses containing human-annotated reasoning errors ("True Mistakes"). **(B)** Heatmaps display the correspondence between human gold-standard annotations (x-axis: x = mistake vs. ✓ = no mistake) and auto-evaluator accuracy (y-axis: success vs. failure). Cell shading reflects row-normalized proportions, and each cell is annotated with the proportion (P) and count (N) of responses; chi-square test results are reported in panel titles. **(C)** Bar plots show the distribution of evaluator-provided reasons among misclassified responses within the "True Mistakes" subset. Reasons were grouped by category (e.g., false negatives, false positives, disagreement) and plotted as the proportion of all misclassifications. Each column represents one model.

We next tested a decomposed prompting strategy in which each potential error subtype was queried independently (Mistake 1→ Mistake 2 …. → Mistake n), allowing the LLM to self-arbitrate. Across models, overall accuracy in detecting and classifying errors remained similar (Figure 6A). Claude-3.5-Sonnet demonstrated a meaningful improvement compared to the previous strategy, increasing agreement with gold-standard labels to 44% (versus 26% in Figure 5) (Figure 6B). Gemini and GPT-4 had more false negatives (under-classification) and persistent false positives (over-classification) (Figure 6C). Claude's distribution of failure modes was relatively unchanged. Although decomposed prompting moderately improved categorization in Claude, it did not uniformly enhance our preceding results.

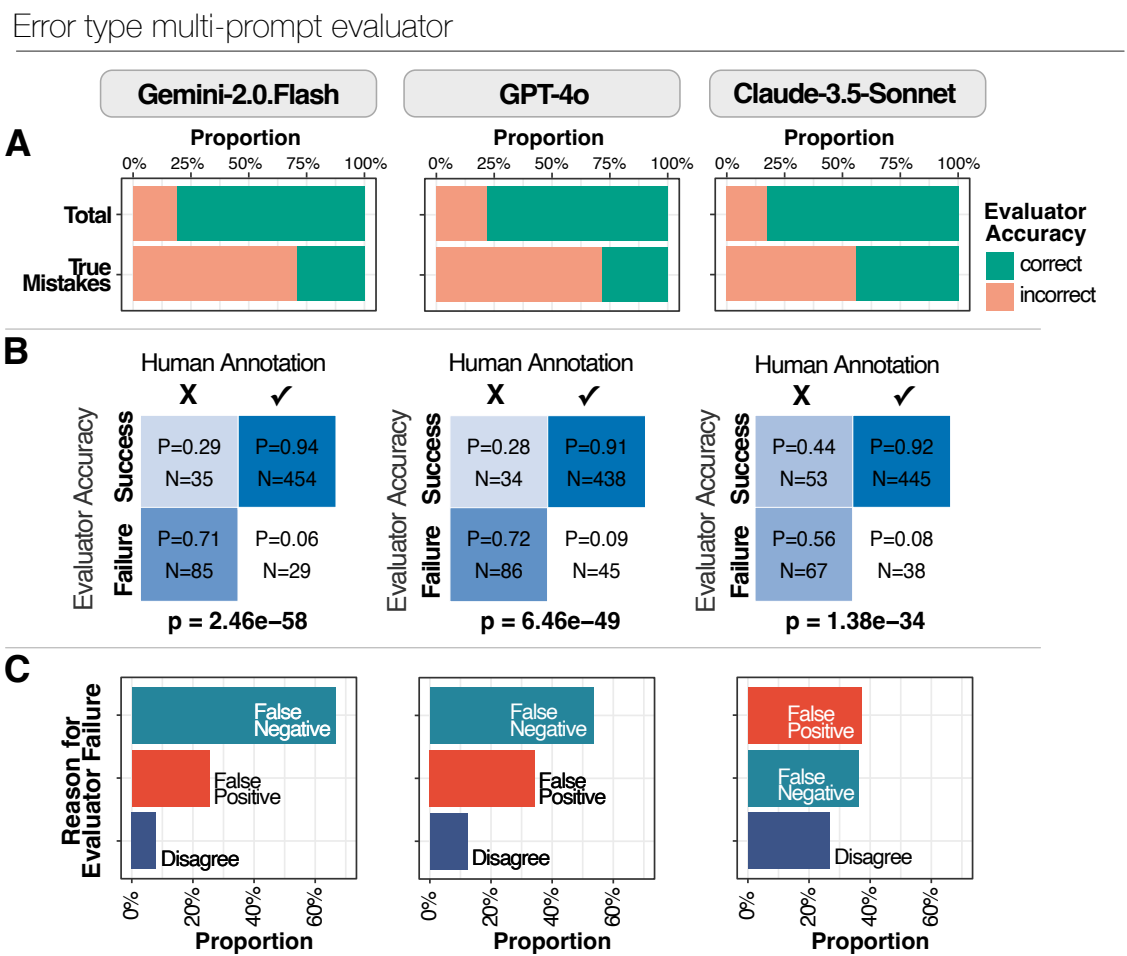

**Figure 6. Decomposed prompting with self-arbitration for automated evaluation.** Models were prompted to evaluate each possible error subtype sequentially (Mistake 1 → Mistake 2 … → Mistake *n*), with outputs compared against human gold-standard annotations from the CORAL oncology dataset. **(A)** Stacked bar plots show the proportion of responses

classified as correct or incorrect, stratified by "All Responses" versus the subset of responses with human-annotated reasoning errors. **(B)** Heatmaps display the correspondence between human annotations (x-axis: x = mistake vs. ✓ = no mistake) and auto-evaluator accuracy to that gold-standard. Row-normalized proportions are indicated by shading, with each tile labeled by proportion (P) and count (N); chi-square test results are reported in panel titles. **(C)** Bar plots show the relative distribution of evaluator-provided reasons among misclassified responses within the "True Mistakes" subset, grouped by reason category. Each column represents one model.

## DISCUSSION

GPT-4 committed reasoning errors linked to cognitive biases in 19.7% of oncology note interpretations. Among recommendations containing reasoning errors, clinical impact scores were significantly lower, with confirmation bias, anchoring bias, and stark omission disproportionately represented among unsafe outputs. This association between reasoning fidelity and clinical safety represents a fundamental gap in current LLM evaluation paradigms, which prioritize endpoint accuracy over logical validity of intermediate reasoning steps[40,41].

The reproducible hierarchical taxonomy we developed reveals reasoning failures could explain 85.4% of mistakes. This three-tier framework, applied across multiple cancer types, provides a clinically conscious approach to classify LLM reasoning errors. By mapping computational failures onto established cognitive bias frameworks, our taxonomy bridges computer science error analysis and clinical reasoning literature[42,43]. Confirmation bias dominated extraction and analysis tasks (31.6% of errors), wherein GPT-4 selectively attended to data aligning with preliminary impressions while dismissing contradictory evidence. Anchoring bias was most prevalent in recommendation tasks, where the model over-weighted initial clinical findings. This task-dependent error distribution implies different contexts elicit distinct failure modes: extraction triggers pattern-matching errors, while recommendation tasks requiring weighing uncertainty expose insufficient deliberation[44–46]. Increased error rates in advanced disease states amplify this risk; the model performs worst precisely when clinical stakes are high and complex.

These findings expose a limitation of accuracy-centric benchmarks, like USMLE and AIME[47,48], where models can reach correct answers through faulty reasoning. In controlled settings this distinction may appear academic[49], yet in clinical practice it becomes critical. Clinical notes are nonlinear documents where contextual cues, provisional impressions, and objective data coexist in complex relationships[50]. Reasoning through these requires causal inference, temporal sequencing, and some skepticism, capabilities that endpoint accuracy alone cannot capture[51,52]. Our taxonomy-based approach exposes subtle reasoning failures missed by accuracy metrics and is scalable to other domains involving complex clinical reasoning.

Perhaps our most actionable finding is the strong association between reasoning errors and guideline discordance. Prospective monitoring of LLM reasoning quality could serve as an early warning system for unsafe recommendations. Error types most strongly associated with discordance—confirmation bias, anchoring bias, and stark omissions—represent targetable failure modes that could be mitigated through adversarial training data[53], debiasing prompts requiring counter-evidence generation[54–56], or risk-stratified oversight where management-phase queries in advanced disease warrant mandatory human review[57,58].

While models like Claude, GPT-4o, and Gemini demonstrated reasonable sensitivity for detecting error presence, they performed poorly at classifying subtypes. This distinction is critical since not all reasoning failures carry equal clinical risk[17,59], and suggests we cannot use isolated models for clinical deployment. Instead, a pragmatic framework resembles distributed quality assurance: automated evaluators perform initial screening while human experts conduct targeted review of flagged cases[60–63]. Such "reasoning audits" embedded in tumor boards or decision-support workflows would protect patients while improving future models. Regulatory frameworks should formalize this human-in-the-loop oversight as a deployment prerequisite.

Study limitations include focus on a single model (GPT-4-32k), though this allowed in-depth characterization of one ubiquitous system. Assessment of clinical harm relied on expert judgment using NCCN guidelines because this was a retrospective, note-based simulation upstream of actual care. We applied zero-shot prompting reflecting realistic usage, though engineered prompts might reduce errors, itself a concern if safe deployment requires specialized expertise. Finally, we did not compare LLM reasoning errors directly against human clinician errors on identical cases, limiting ability to contextualize errors.

As LLMs move from experimental tools to clinical decision-support systems, their value will depend on reasoning soundness, not output fluency. We show GPT-4, despite high performance on benchmarks, commits reasoning errors correlating with guideline-discordant and potentially harmful recommendations. The hierarchical taxonomy we developed offers a diagnostic schema for identifying these failures and possibly transferable across clinical domains. Ensuring clinical deployment requires models reach correct conclusions through logically valid and clinically defensible reasoning. Anything less risks encoding the very cognitive biases causing human clinicians to err, transforming a tool meant to augment judgment into one that undermines it.

## 1. Development Phase

**Prompt development**
- Clinical **zero-shot** tasks for extraction, analysis, and recommendation + chain of though prompting

*4 iterations of prompt engineering*

**Annotation and post-hoc analysis:**
- Generate hierarchy of chain-of-thought reasoning with **GPT4-32k**

*Reviewed 150 CoT responses from 15 prompts across 5 breast and 5 pancreatic cancer notes*

## 2. Validation Phase

**Cohort:**
**40 Patients (CORAL)**
- 20 breast cancer cases
- 20 pancreatic adenocarcinoma cases
- All progress notes

**+**

**Tasks:**
- **Extraction (n=15)** *Questions focused on extracting and analyzing clinical information found in most oncology notes.*   *(Table S2A)*

**=**

**600 GPT4 responses** generated and manually annotated by **error taxonomy**

## 3. Testing Phase

**Cohort:**
**24 Patients (Mayo)**
- 6 patients per prostate cancer stage (local, biochemical recurrence, MCRPC, CSPC)
- 1 note / patient *(exception: additional RadOnc or Urology note included for 3/6 of localized PCa patients)*
- Assessment/Plan masked from input

**+**

**Tasks:**
- **Extraction (n=15)** *Questions focused on extracting and analyzing clinical information found in most oncology notes.*
- **PCa stage-specific:** *NCCN PCa guideline-based, clinical analysis and recommendation tasks [localized (10), biochemical recurrence (5), MCRPC (5), CSPC (5)]*   *(Table S3A)*

**=**

**822 GPT4 responses** generated and manually annotated by **error taxonomy**, **clinical impact**, and **guideline concordance**

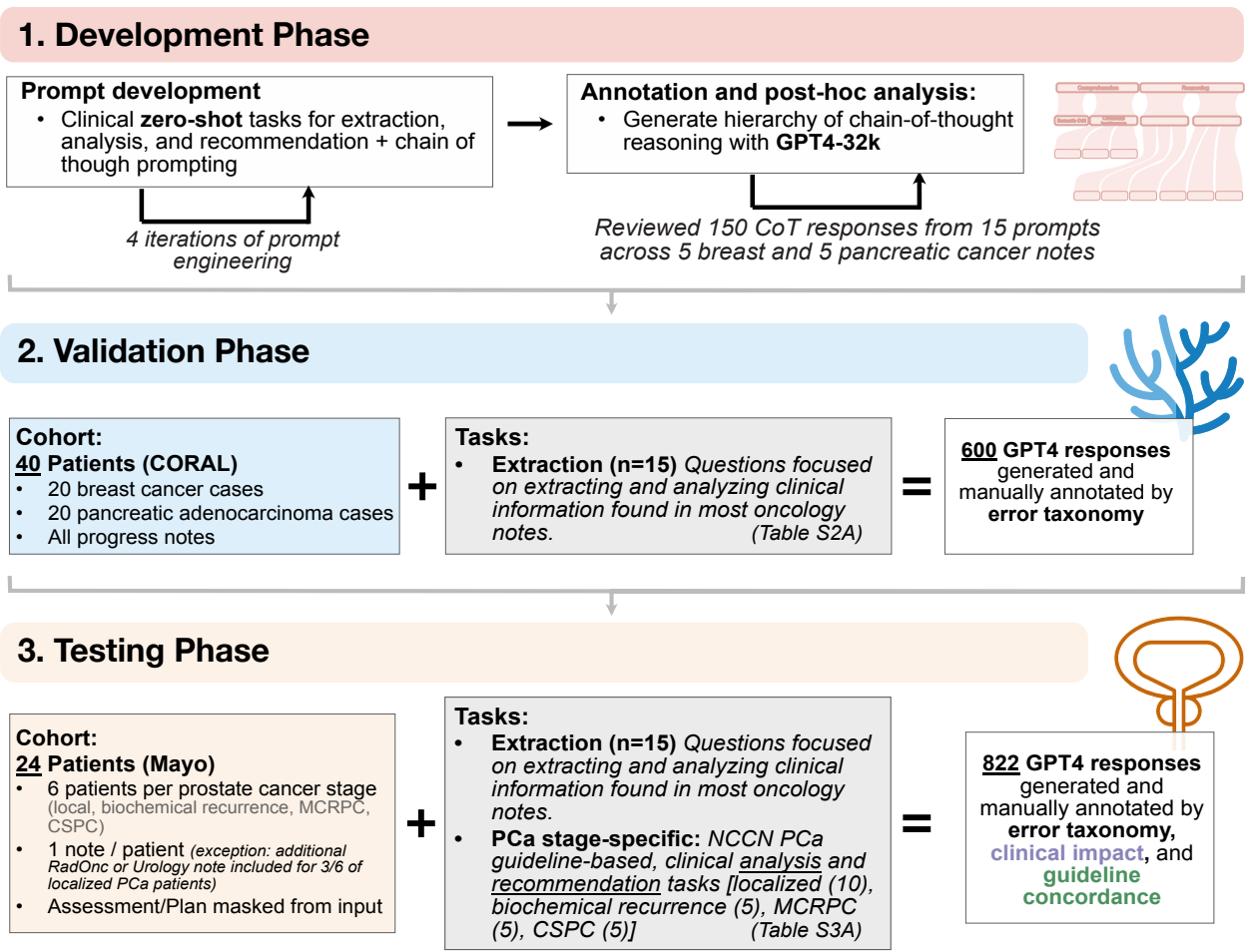

**Supplementary Figure 1.** Framework for Annotating Errors in LLM Responses.